\documentclass{article}
\PassOptionsToPackage{numbers, compress}{natbib}
\usepackage[nonatbib, preprint]{neurips_2023}

\usepackage{amsmath,amsfonts,bm}

\def\eqref#1{equation~\ref{#1}}

\def\1{\bm{1}}

\def\rvb{{\mathbf{b}}}
\def\rvc{{\mathbf{c}}}

\def\rvm{{\mathbf{m}}}

\def\rvo{{\mathbf{o}}}

\def\rvq{{\mathbf{q}}}

\DeclareMathAlphabet{\mathsfit}{\encodingdefault}{\sfdefault}{m}{sl}
\SetMathAlphabet{\mathsfit}{bold}{\encodingdefault}{\sfdefault}{bx}{n}

\usepackage{url}
\usepackage[utf8]{inputenc} 
\usepackage[T1]{fontenc}    
\usepackage{url}            
\usepackage{booktabs}       
\usepackage{amsfonts}       
\usepackage{nicefrac}       
\usepackage{microtype}      
\usepackage{colortbl}
\usepackage[normalem]{ulem}
\usepackage{enumitem}
\usepackage{algorithmic}
\usepackage[linesnumbered,ruled,vlined]{algorithm2e}
\usepackage{hhline}
\usepackage{times}
\usepackage{epsfig}
\usepackage{graphicx}
\usepackage{amsmath}
\usepackage{amssymb}
\usepackage{multirow}
\usepackage{subcaption}
\usepackage{pifont}%
\usepackage{wrapfig}
\newcommand{\cmark}{\ding{51}}%
\newcommand{\xmark}{\ding{55}}%
\SetKwComment{Comment}{\color{green!50!black}\# }{}

\SetKwProg{Function}{def}{:}{}

\SetKwProg{For}{for}{:}{}
\SetKwProg{If}{if}{:}{}

\definecolor{myblue}{RGB}{88, 190, 237}
\definecolor{deemph}{gray}{0.6}

\usepackage[breaklinks=true,bookmarks=false]{hyperref}

\title{Semantic-SAM: Segment and Recognize Anything at Any Granularity}

\author{
  ~Feng Li$^{\spadesuit*}$, ~~Hao Zhang$^{\spadesuit*}$, ~~Peize Sun$^{\sharp}$, ~~Xueyan Zou$^{\S}$,~~Shilong Liu$^{\P}$, ~~Chunyuan Li$^{\ddagger}$ \\
\textbf{
~Jianwei Yang$^{\ddagger1}$,~Lei Zhang$^{\dagger2}$, ~Jianfeng Gao$^{\ddagger2}$
}
\and
{
\footnotesize
$^{\spadesuit}$ HKUST \;
$^\ddagger$ Microsoft Research, Redmond \;  
$^\dagger$ IDEA \;  
$^{\sharp}$ HKU \;
$^{\S}$ UW-Madison \;
$^{\P}$ Tsinghua \;
}
\and
\scriptsize{
$^*$~Equal Contribution \;
$1.$~Project Lead \;
$2.$~Equal Advisory Contribution \;
}
}

\newcommand{\ourmodel}{Semantic-SAM}

\begin{document}

\maketitle

\begin{abstract}

In this paper, we introduce \textit{\ourmodel}, a universal image segmentation model to enable segment and recognize anything at any desired granularity. Our model offers two key advantages: semantic-awareness and granularity-abundance. To achieve semantic-awareness, we consolidate multiple datasets across granularities and train on decoupled objects and parts classification. This allows our model to facilitate knowledge transfer among rich semantic information. For the multi-granularity capability, we propose a multi-choice learning scheme, enabling each click point to generate masks at multiple levels that correspond to multiple ground-truth masks. Notably, this work represents the first attempt to jointly train a model on SA-1B, generic, and part segmentation datasets. Experimental results and visualizations demonstrate that our model successfully achieves semantic-awareness and granularity-abundance. Furthermore, combining SA-1B training with other segmentation tasks, such as panoptic and part segmentation, leads to performance improvements. We will provide code and a demo for further exploration and evaluation at \url{https://github.com/UX-Decoder/Semantic-SAM}.

\end{abstract}

\section{Introduction}

The universal and interactive AI systems that follow human intents have shown their potential in natural language processing~\cite{openai2022chatgpt,openai2023gpt4} and controllable image generation~\cite{rombach2022high,zhang2023adding}. However, such a universal system for pixel-level image understanding remains less explored. 
We argue that a universal segmentation model should possess the following important properties: \textit{universal representation},
\textit{semantic-awareness}, and \textit{granularity-abundance}. Regardless of the specific image domain or prompt context, the model is capable of acquiring a versatile representation, predicting segmentation masks in multi-granularity, and understanding the semantic meaning behind each segmented region.

Previous works~\cite{kirillov2023segment,zou2023segment,wang2023seggpt} attempted to investigate these properties, but only achieved part of the goals. The main obstacles impeding the progress of such a universal image segmentation model can be attributed to limitations in both model architecture flexibility and training data availability.

\begin{itemize}[leftmargin=*]
\item \textbf{Model Architecture.} The existing image segmentation model architectures are dominated by the single-input-single-output pipeline that discards any ambiguity. While this pipeline is prevalent in both anchor-based CNN architectures~\cite{he2017mask} and query-based Transformer architectures~\cite{carion2020end,cheng2022masked}, and has demonstrated remarkable performance in semantic, instance, and panoptic segmentation tasks~\cite{lin2014microsoft,zhou2018semantic,kirillov2019panoptic}, it inherently restricts the model to predict multi-granularity segmentation masks in an end-to-end manner. Although clustering postprocessing techniques~\cite{de2021part} can produce multiple masks for a single object query, they are neither efficient nor effective solutions for a granularity-aware segmentation model.

\item \textbf{Training Data.} Scaling up segmentation datasets that possess both semantic-awareness and granularity-awareness is a costly endeavor. Existing generic object and segmentation datasets such as MSCOCO~\cite{lin2014microsoft} and Objects365~\cite{shao2019objects365} offer large amounts of data and rich semantic information, but only at the object level. On the other hand, part segmentation datasets such as Pascal Part~\cite{chen2014detect}, PartImageNet~\cite{he2021partimagenet}, and PACO~\cite{ramanathan2023paco} provide more fine-grained semantic annotations, but their data volumes are limited. Recently, SAM~\cite{kirillov2023segment} has successfully scale up the multi-granularity mask data to millions of images, but it does not include semantic annotations. In order to achieve the dual objectives of semantic-awareness and granularity-abundance, there is a pressing need to unify segmentation training on various data formats to facilitate knowledge transfer. However, the inherent differences in semantics and granularity across different datasets pose a significant challenge to joint training efforts.
   
\end{itemize}

\begin{figure}[t!]
    \centering
    \includegraphics[width=0.97\textwidth]{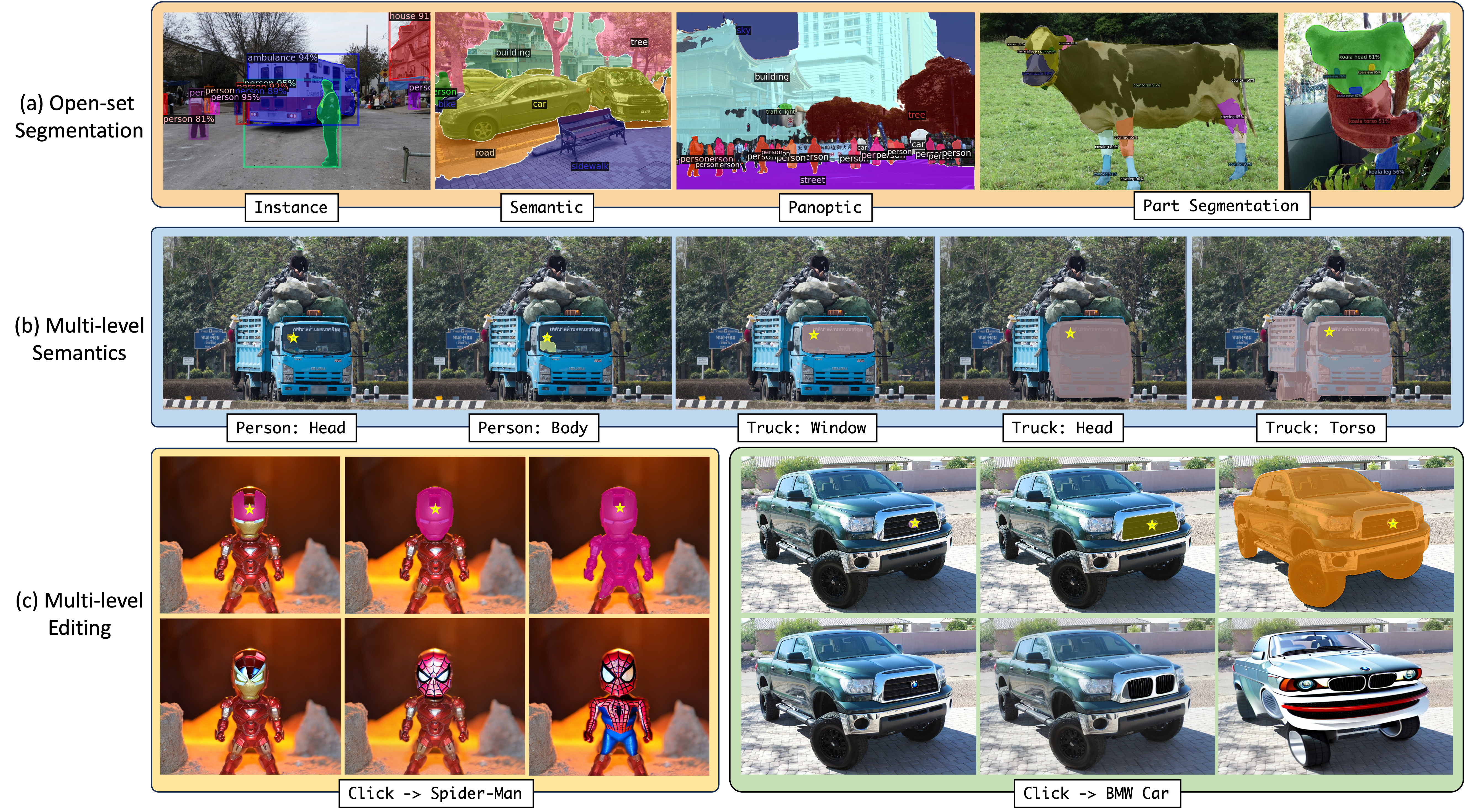}
    \vspace{-5pt}
    \caption{Our model is capable of dealing with various segmentation tasks including open-set and interactive segmentation. (a) Our model can do instance, semantic, panoptic segmentation, and part segmentation. (b) Our model is able to output multi-level semantics with different granularities. The red point on the left-most image is the click.(c) We connect our model with an inpainting model to perform multi-level inpainting. The prompts are "Spider-Man" and "BMW car", respectively. Note that only one click is needed to produce the results in (b) and (c), respectively.}
    \label{fig:teaser}
    \vspace{-12pt}
\end{figure}
In this paper, we introduce \textit{Semantic-SAM}, a universal image segmentation model designed to enable segmenting and recognizing objects at any desired granularity. Given one click point from a user, our model addresses the spatial ambiguity by predicting masks in multiple granularities, accompanied by semantic labels at both the object and part levels. As shown in Figure~\ref{fig:teaser}, our model generates multi-level segmentation masks ranging from the person head to the whole truck.

The multi-granularity capability is achieved through a multi-choice learning design~\cite{li2018interactive,guzman2012multiple} incorporated into the decoder architecture. Each click is represented with multiple queries, each containing a different level of embedding. These queries are trained to learn from all available ground-truth masks representing different granularities. To establish a correspondence between multiple masks and ground-truths, we employ a many-to-many matching scheme to ensure that a single click point could generate high-quality masks in multiple granularities.

To accomplish semantic-awareness with a generalized capability, we introduce a decoupled classification approach for objects and parts, leveraging a shared text encoder to encode both objects and parts independently. This allows us to perform object and part segmentation separately, while adapting the loss function based on the data type. For instance, generic segmentation data lacks part classification loss, whereas SAM data does not include classification loss.

To enrich semantics and granularity within our model, we consolidate seven datasets on three types of granularities, including generic segmentation of MSCOCO~\cite{lin2014microsoft}, Objects365~\cite{shao2019objects365}, ADE20k~\cite{zhou2018semantic}, part segmentation of PASCAL Part~\cite{chen2014detect}, PACO~\cite{ramanathan2023paco}, PartImagenet~\cite{he2021partimagenet}, and SA-1B~\cite{kirillov2023segment}. Their data formats are reorganized to match our training objectives accordingly. After joint training, our model obtains a strong performance across a variety of datasets. Notably, we find that learning from interactive segmentation could improve generic and part segmentation. For example, by jointly training SA-1B promptable segmentation and COCO panoptic segmentation, we achieve a gain of \textbf{2.3} box AP and a gain of \textbf{1.2} mask AP. In addition, through comprehensive experiments, we demonstrate that our granularity completeness is better than SAM with more than 3.4 1-IoU.

\section{Data Unification: Semantics and Granularity}
In order for multi-level semantics, we include seven datasets that contain different granularity-level masks. The datasets are SA-1B, COCO panoptic, ADE20k panoptic, PASCAL part, PACO, PartImageNet, and Objects365. Within them, COCO and ADE20k panoptic datasets contain object-level masks and class labels. PASCAL part, PACO, and PartImageNet contain part-level masks and class labels. SA-1B contains up to 6-level masks without labels, while Objects365 contains abundant class labels for object-level instances. The details of these datasets are shown in Table~\ref{tab:data_used}. We further visualize the data distribution of different data type in Fig~\ref{fig:data_used_chart}.

\begin{minipage}{\textwidth}
  \begin{minipage}[b]{0.56\textwidth}
    \centering
    \addtolength{\extrarowheight}{\belowrulesep}
    \resizebox{0.99\linewidth}{!}{\begin{tabular}{lcc|cc|cc} 
    \toprule
    \multirow{2}{*}{Type}&\multirow{2}{*}{Data}  &\multirow{2}{*}{\#Images}& \multicolumn{2}{c|}{Semantic Concept} & \multicolumn{2}{c}{Granularity Level}     \\
     &&& Part  & Object & Part & Whole   \\ 
     \hline
    Class-agnostic & SA-1B   &11B& \xmark &  \xmark & \cmark &\cmark \\
     \hline
     \multirow{2}{*}{Object-level}&
     Objects365&1.7M   & \xmark &  365 & \xmark &\cmark  \\
     &COCO&110K   &  \xmark& 133  & \xmark &\cmark \\
     &ADE20K&20K  &  \xmark & 150  & \xmark &\cmark \\
       \hline
      \multirow{3}{*}{Part-level}&
      PACO-LVIS&45K    &  201& 75 & \cmark &\cmark \\
      &PartImageNet&16K   & 13 &  11 & \cmark &\cmark \\
      &Pascal Part&5K   & 30 &  20 & \cmark &\cmark \\
    \bottomrule
    \end{tabular}}
      \captionof{table}{The data statistics  in \ourmodel{}.}
      \label{tab:data_used}
    \end{minipage}
\hfill
  \begin{minipage}[b]{0.43\textwidth}
    \centering
    \includegraphics[width=0.97\textwidth]{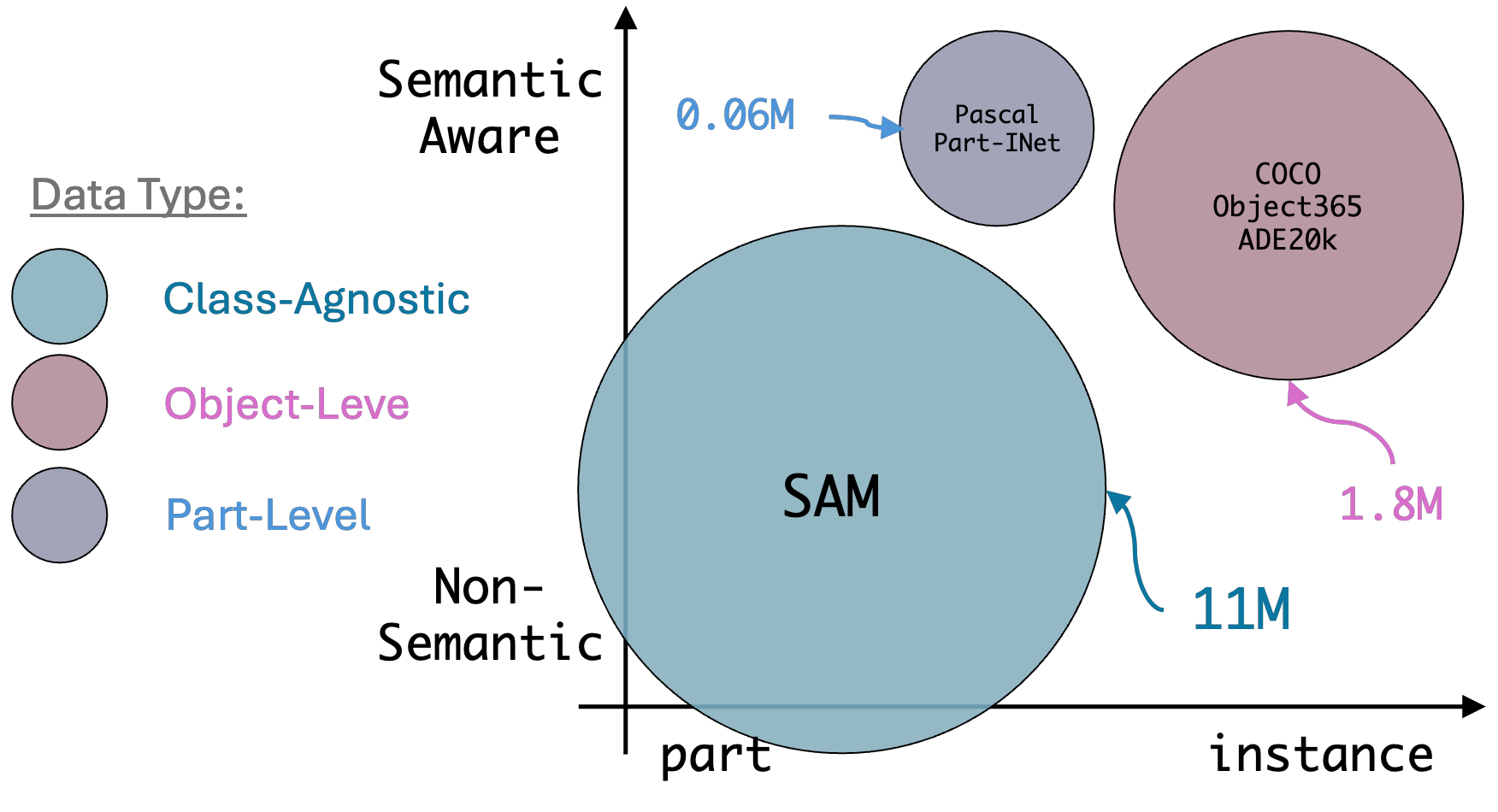}
    \captionof{figure}{Semantics-Granularity 2D chart.}
    \label{fig:data_used_chart}
  \end{minipage}    
  \end{minipage}

\section{\ourmodel{}}
\subsection{Model}
\begin{figure}[t!]
    \centering
    \includegraphics[width=0.97\textwidth]{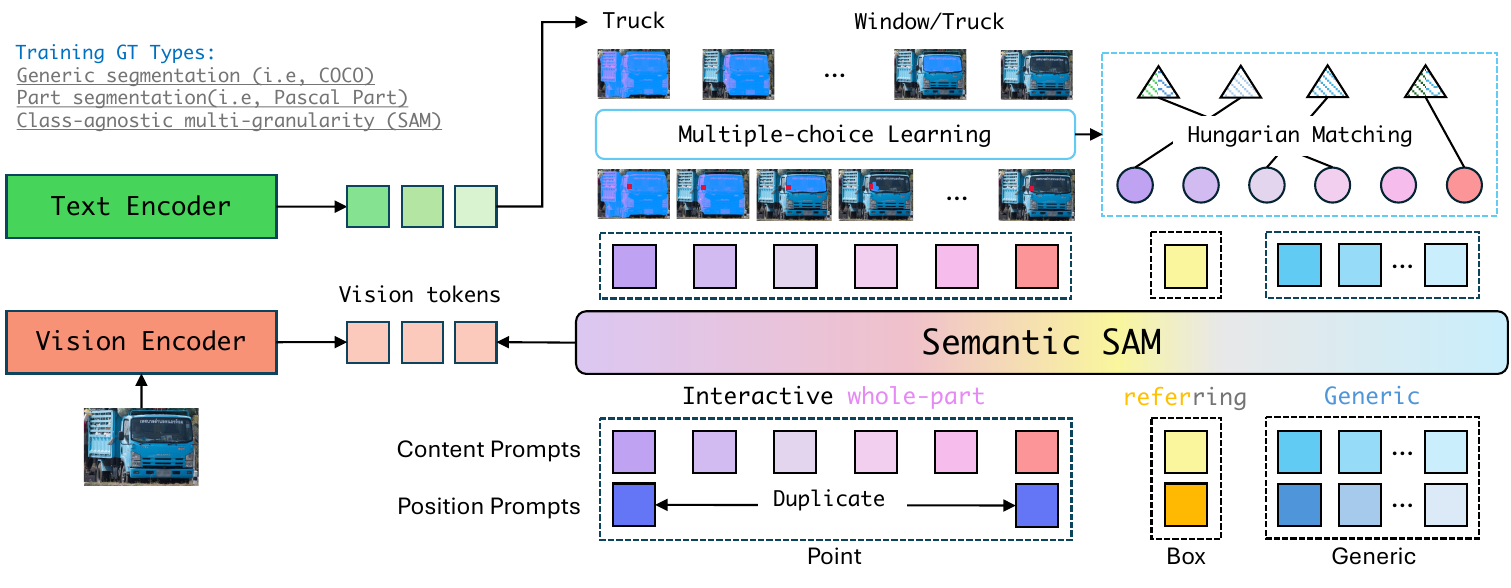}
    \vspace{-5pt}
    \caption{\ourmodel{} is a universal segmentation framework that can take multiple types of segmentation data including generic, part, and class-agnostic segmentation data. The Vision Encoder is used to extract image features. The mask decoder can do both generic segmentation and promptable segmentation with various types of prompts. For point and box, we input them via anchor boxes to the mask decoder. 
    Since there is an ambiguity of granularity for a point input, we duplicate each point $6$ times and give them different levels of embeddings. The output masks of point prompts match with multiple GT masks of different granularities. }
    \label{fig:model_overview}
    \vspace{-12pt}
\end{figure}

Our \ourmodel{} follows \cite{li2022mask} to exploit a query-based mask decoder to produce semantic-aware and multi-granularity masks. In addition to the generic queries, it supports two types of prompts including point and box, similar to SAM~\cite{kirillov2023segment}. The overall pipeline is shown in Fig.~\ref{fig:model_overview}. 

We represent both click and box prompts into anchor boxes as a unified format. In particular, we convert user click point $(x, y)$ into an anchor box $(x, y, w, h)$ with small width $w$ and height $h$, so that the anchor box can closely approximate the point. To capture different granularities of masks, each click is first encoded to position prompt and combined with $K$ different content prompts, where each content prompt is represented as a trainable embedding vector for a given granularity level. Here we empirically choose $K=6$, considering there are at most 6 levels of masks per user click for the majority of images in SA-1B~\cite{kirillov2023segment}. 
More specifically, a click/box $\mathbf{b}=(x, y, w, h)$ is encoded into $K$ content embeddings and one position embedding, respectively. We represent its content embeddings as a set of query vectors $\mathbf{Q} = ( \rvq_1, \cdots, \rvq_K)$. For the $i$-th query,  
\begin{equation}
    \mathbf{q}_i=\mathbf{q}^{\texttt{level}}_i+\mathbf{q}^{\texttt{type}}_{i}, 
    \label{eq:click_encoding}
\end{equation}
where
\begin{itemize}[leftmargin=7.5mm]
\setlength{\itemsep}{2pt}
\item 
$\mathbf{q}^{\texttt{level}}$ is the embedding for granularity level $i$,
\item
$\mathbf{q}^{\texttt{type}}$ distinguishes the query type, chosen from either the click or the box embeddings.
\end{itemize}

The position embedding of $\mathbf{c}$ is implemented via sine encoding. Assuming that the output image feature from vision encoder is $\mathbf{F}$, the mask decoder of the proposed \ourmodel{} represents the click on the input image as:
\begin{align}
 \mathbf{O}= \texttt{DeformDec} (\mathbf{Q},\rvb, \mathbf{F} ) ~ \text{with}~ 
    \textbf{O}=(\rvo_1, \cdots, \rvo_K),
    \label{eq:model_output}
\end{align}
where $\texttt{DeformDec} (\cdot,\cdot,\cdot)$ is a deformable decoder that takes query feature, reference box, and image features as input to output queried features. $\rvo_i$ is the model output for the $i$th input query $\rvq_i$. Each $\rvo_i=(\rvc_i, \rvm_i)$ consists of the predicted semantic category $\rvc_i$ and mask $\rvm_i$, which are used to construct the concept recognition loss and mask prediction loss, respectively.

\subsection{Training}

\begin{wrapfigure}{r}{0.53\textwidth}
  \begin{minipage}{0.53\textwidth}
\centering  
\vspace{-10mm}
\includegraphics[width=0.90\textwidth]{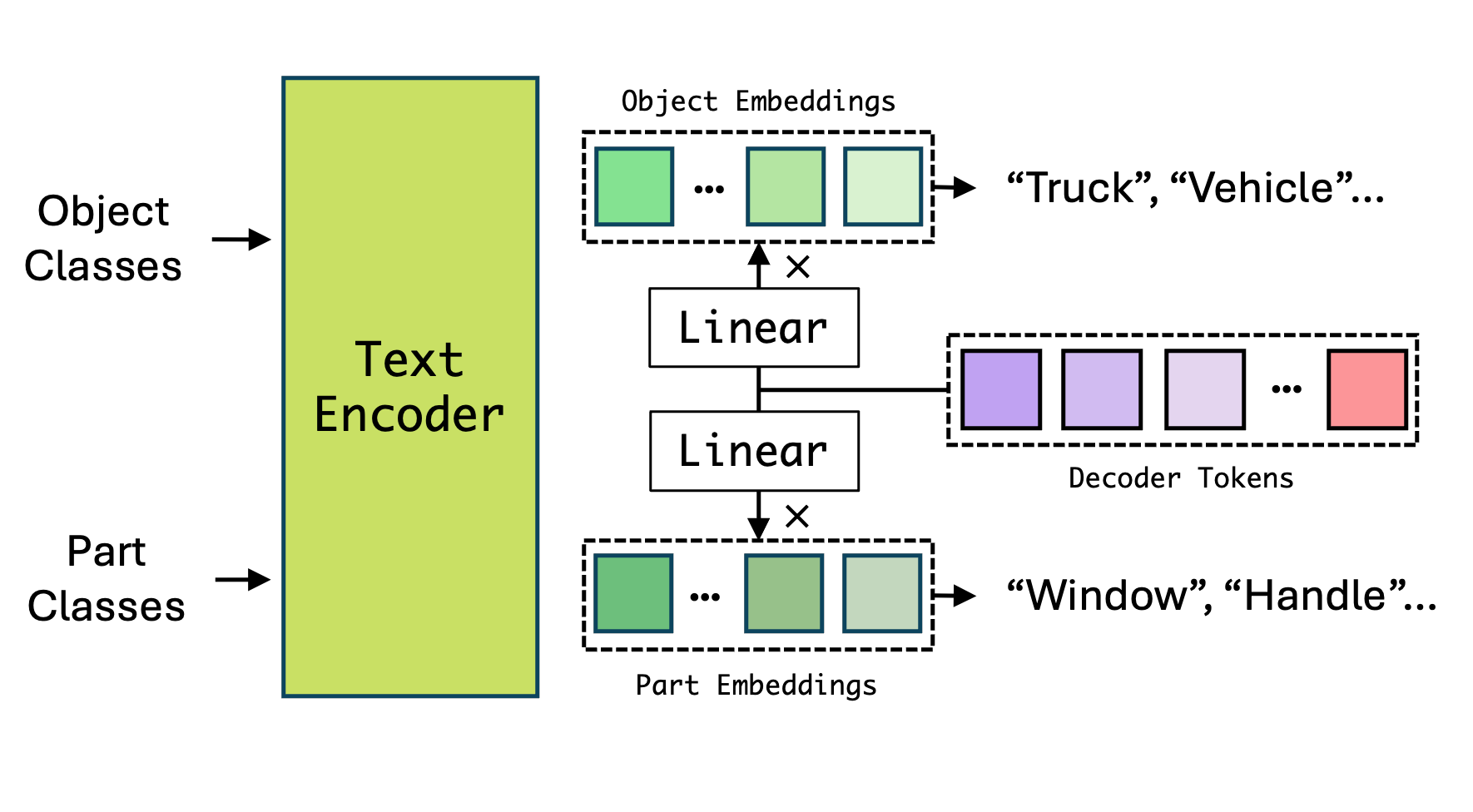}
\vspace{-7mm}
\captionof{figure}{Decoupled object and part classification.}  
\label{fig:classification}
\vspace{-10pt}
  \end{minipage}
\end{wrapfigure}
\paragraph{Recognize Anything.}
As we train with various types of data with different semantic annotations, in which some contain object-level annotations (COCO), some contain both object and part-level annotations (Pascal Part), and SA-1B has no semantic annotations but contains masks of all semantic levels. Note that a large number of part concepts are shared across different objects, for example, \textit{head} for all animals. We aim to transfer the part concept knowledge across objects trained with only object-level annotations in our joint training.
To address this discrepancy between semantic annotations and better transfer semantics of different granularity, we propose to decouple object and part recognition. As shown in Fig~\ref{fig:classification}, we utilize a shared text encoder to encode objects and parts, which are used to perform object and part segmentation separately. 

Importantly, while all types of segmentation data share a unified format, the loss varies for different data types. We summarize the loss items to construct the training objective in \ourmodel{} in Table~\ref{tab:loss_items}. It is the part-level data that bridges the gap to recognize semantic concepts between part and object levels, and it is the use of SAM data in Hungarian matching that bridges the gap to segment masks at any granularity.

\begin{table*}[h!]
\centering
\addtolength{\extrarowheight}{\belowrulesep}
\resizebox{0.49\linewidth}{!}{\begin{tabular}{l|cc|ccc} 
\toprule
\multirow{2}{*}{Data}  & \multicolumn{2}{c|}{Recognize Anything} & \multicolumn{3}{c}{Segment at Any Granularity}     \\
 & Part  & Object & Box & Mask & \#GT in Matching  \\ 
 \hline
 SAM data   & \xmark &  \xmark & \cmark &\cmark & Many  \\
 Object-level data   & \xmark &  \cmark & \cmark &\cmark & One  \\
 Part-level data   & \cmark &  \cmark & \cmark &\cmark & One  \\
\bottomrule
\end{tabular}}
\caption{The loss items to construct the training objective in \ourmodel{}. The four loss items are part classification, object classification, box loss and mask loss, respectively. The last column indicates the number of ground-truth mask in the matching.}
\label{tab:loss_items}
\vspace{0.2cm}
\end{table*}

\paragraph{Segment at any granularity.}
To endow the model with a multi-granularity segmentation ability, we propose a many-to-many matching method during training. We found that SAM fails in providing good multi-level segmentation results with a single click because SAM uses many-to-one matching during training. In other words, the three SAM-predicted masks for each click only match with one GT mask. This causes that points located in masks of small levels cannot predict large masks with high quality according to our observation. In contrast, to enable multi-level mask prediction with a single click, we fully leverage the structures in both data and algorithm. First, we re-organize the data by clustering multiple GT masks of different levels sharing the same click. To allow multiple predictions of the same click to match with the GT masks, we employ the Hungarian algorithm to enable the many-to-many matching. The similarity matrix and scores vary based on the availability of different segmentation data components.

For box input and generic segmentation, we follow existing methods. Specifically, to generate a mask from an input box, we follow a similar idea as in denoising training (DN)~\cite{li2022mask}. We add noises to ground-truth boxes to simulate inaccurate box inputs from users, and these noised boxes serve as spatial prompts for the decoder. The model is trained to reconstruct the original boxes and masks given noised boxes. For the content part of box prompts, we adopt a learnable token as a general prompt. Note that this is the only difference from DN, as DN uses ground-truth label embedding as the content prompts.
For generic segmentation, we follow the same pipeline as in Mask DINO~\cite{li2022mask}.

\begin{figure}[t]
    \centering
    \includegraphics[width=0.88\textwidth]{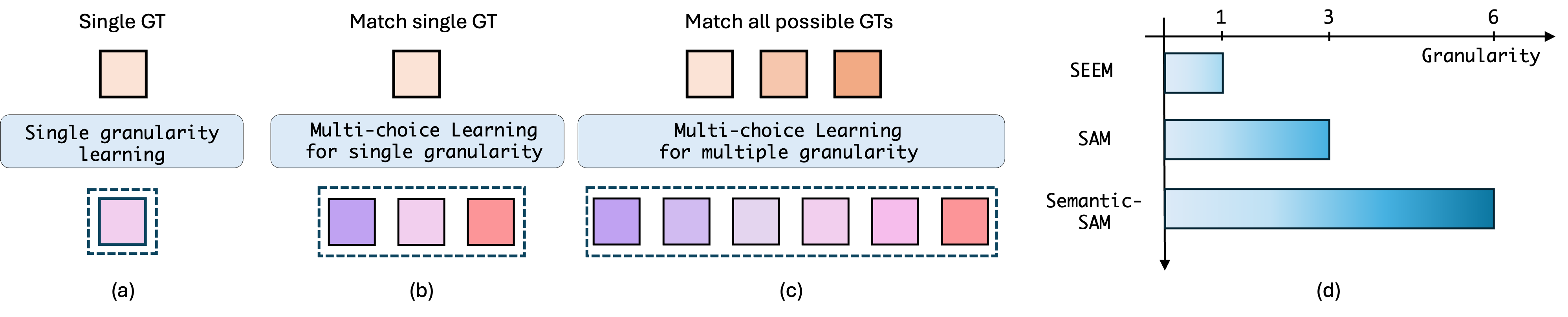}
    \vspace{-5pt}
    \caption{Inteactive learning strategy comparison between a) One-to-one: traditional interactive segmentation models that focus on object-level, i.e, SEEM, b) Many-to-one: multi-choice learning for single granularity, i.e, SAM, and c) Many-to-many: ours. We enforce the model to predict all the possible granularities of a single click for more controllable segmentation. d) As a result, our output granularity are richer to generate diverse output masks.
    }
    \label{fig:match}
\end{figure}

\paragraph{Discussion.} 
As shown in Fig.~\ref{fig:match}, compared with previous interactive segmentation models, \ourmodel{} differs from previous segmentation models in two aspects.  Firstly, we train the model to output all the possible segmentation masks with one click. Secondly, our output granularities are richer to generate diverse output masks.

\section{Experiments}
\subsection{Experimental Setup}
\noindent

\noindent
\textbf{Implementation Details.}
In our experiments, we jointly train on three types of data, as shown in Table~\ref{tab:data_used}.
We implement our model based on Mask DINO~\cite{li2022mask} . Mask DINO is a unified detection and segmentation framework which simultaneously predicts box and mask. We follow~\cite{li2022mask} to use 300 latent queries and nine decoder layers for all segmentation tasks. For the visual backbone, we adopt pre-trained Swin-T/L~\cite{liu2021swin} by default. For the language backbone, we adopt the pre-trained base model in UniCL~\cite{yang2022unified}. 
As SA-1B~\cite{kirillov2023segment} dominates the data, during training, we first train on only SA-1B data. Then, we add object and part-level data to jointly train the three types of data. During training, the image resolution is $1024\times 1024$ for all data. We use AdamW~\cite{loshchilov2017decoupled} as the optimizer. We use large-scale jittering for object and part-level data and did not use data augmentations for SA-1B data, as SA-1B images are abundant. We set the learning rate to $0.0001$, which is decayed at 0.9 and 0.95 fractions of the total number of steps by 10.
\smallskip

\noindent
\textbf{Evaluation.} We mainly evaluate two datasets, including COCO \textit{Val2017} and a subset of SA-1B~\cite{kirillov2023segment} with 1000 images. For evaluation metrics, we evaluate PQ and AP for generic and part segmentation datasets. For single-granularity interactive segmentation, we report Point (Max) and Point (Oracle). \textit{Max} denotes we select the output mask with the maximum confidence score. \textit{Oracle}
denotes we select the output mask with the max IoU by calculating the IoU between the prediction and target mask. For multi-granularity interactive segmentation, we report 1-IoU@All Granularity that matches all the possible ground-truth masks for a single click to the multi-granularity predictions and then calculate the average IoU of all granularities.
\begin{table*}
\centering
\addtolength{\extrarowheight}{\belowrulesep}
\footnotesize
\resizebox{0.99\linewidth}{!}{\begin{tabular}{l|c|c|c|c|cc|cc|cc|cc} 
\toprule
\multirow{2}{*}{Method} &\multirow{2}{*}{Type}& \multirow{2}{*}{Training Data}& \multirow{2}{*}{PQ}& \multirow{2}{*}{mIoU} & \multicolumn{2}{c|}{AP} & \multicolumn{2}{c|}{APs} & \multicolumn{2}{c|}{APm} & \multicolumn{2}{c}{APl}    \\
 &  & &&&box & mask&box & mask&box & mask&box & mask   \\ 
 \hline
 Mask2Former (T) \cite{cheng2022masked}   & Close-set& COCO& {53.2} & 63.2 & $46.1$ & $43.3$ & $-$ & $-$& $-$& $-$& $-$& $-$\\
  X-Decoder (T) \cite{zou2022generalized}   & Open-set& COCO+VL& {52.6} & 62.4 & $43.6$ & $41.3$ & $-$ & $-$& $-$& $-$& $-$& $-$ \\
  \hline
 OpenSeed (T) \cite{zhang2023simple}  & Open-set& COCO+O365& {55.4} & 63.8 & $51.2$ & $47.1$ & $34.5$& $27.4$& $54.3$& $50.4$& $66.2$& $66.8$ \\
  \ourmodel{} (T) (ours)   & Open-set& COCO&54.6&63.2&50.1&46.1&34.4&27.1&53.2&49.4&66.1&66.1  \\
         \ourmodel{}  (T) (ours)   & Open-set& COCO+SAM &55.2&63.4&52.3\fontsize{8.0pt}{\baselineskip}\selectfont{(+2.2)}&47.4\fontsize{8.0pt}{\baselineskip}\selectfont{(+1.3)}&36.1\fontsize{8.0pt}{\baselineskip}\selectfont{(+1.7)}&28.3\fontsize{8.0pt}{\baselineskip}\selectfont{(+1.2)}&55.6\fontsize{8.0pt}{\baselineskip}\selectfont{(+2.4)}&50.7\fontsize{8.0pt}{\baselineskip}\selectfont{(+1.3)}&67.3&66.2 \\
\bottomrule
\end{tabular}}
\caption{Results for \ourmodel{} and other panoptic segmentation models on COCO \texttt{val2017}. Our model is jointly trained on COCO~\cite{chen2015microsoftcoco} and ~\cite{kirillov2023segment} (1/10 data) and directly evaluates COCO.}
\label{tab:generic}
\vspace{0.2cm}
\end{table*}

\subsection{Semantic Segmentation of Anything}

\paragraph{Generic Segmentation}

As shown in Table~\ref{tab:generic}, to validate the compatibility of multi-granularity interactive segmentation and generic segmentation, we jointly train with SA-1B~\cite{kirillov2023segment} (1/10 data) and COCO panoptic segmentation. The result indicates that interactive segmentation with SAM can significantly help the instance-level detection and segmentation with a performance improvement of +2.2 AP on the box and +1.3 AP on the mask. Notably, OpenSeed~\cite{zhang2023simple} and \ourmodel{} are both based on Mask DINO~\cite{li2022mask}. Our joint training with SA-1B even outperforms OpenSeed which is trained with Object365~\cite{shao2019objects365}. In addition, adding SA-1B mainly improves small object detection (APs and APm), as there are a large number of small objects in SA-1B.

\paragraph{Part Segmentation}
\begin{table*}
\centering
\addtolength{\extrarowheight}{\belowrulesep}
\footnotesize
\resizebox{0.99\linewidth}{!}{\begin{tabular}{l|c|c|cc|cc|cc|cc} 
\toprule
\multirow{2}{*}{Method} &\multirow{2}{*}{Type}& \multirow{2}{*}{Training Data}& \multicolumn{2}{c|}{AP} & \multicolumn{2}{c|}{APs} & \multicolumn{2}{c|}{APm} & \multicolumn{2}{c}{APl}    \\
 &  &&box & mask&box & mask&box & mask&box & mask   \\ 
 \hline
VLPart~\cite{sun2023going} & Open-set& Pascal Part&$-$&27.4&$-$&$-$&$-$&$-$&$-$&$-$ \\
\ourmodel{} (ours)   & Open-set& Pascal Part& 27.0&30.5&16.6&19.1&38.1&41.6&43.8&49.1\\
         \ourmodel{} (ours)   & Open-set& Pascal Part+SAM & 28.0&31.4&17.3&19.9&40.0&42.5&45.7&49.7  \\

\bottomrule
\end{tabular}}
\caption{Results for \ourmodel{} and other part segmentation models on Pascal Part. Our model is jointly trained on Pascal Part~\cite{everingham2011pascal} and SA-1B~\cite{kirillov2023segment} (1/10 data) and directly evaluates Pascal Part.}
\label{tab:part}
\vspace{0.2cm}
\end{table*}



We also validate the compatibility of joint training SA-1B (1/10 data) and part segmentation. As shown in Table~\ref{tab:part}, adding SA-1B brings a decent performance improvement on Pascal Part~\cite{everingham2011pascal}.


\paragraph{Single-granularity Interactive Segmentation}

\begin{table*}
\centering
\begin{tabular}{l|cc} 
\hline
\multirow{3}{*}{Method} & \multicolumn{2}{c}{COCO}  \\
& Point (Max) & Point (Oracle)   \\
& 1-IoU & 1-IoU \\ 
\hline
SAM (B) &52.1&68.2\\
SAM (L) &55.7&70.5\\

\ourmodel{} (T) &54.5&73.8\\
\ourmodel{} (L) &57.0&74.2 \\
\hline
\end{tabular}
\caption{Comparison with previous models on point interactions. Both SAM~\cite{kirillov2023segment} and our model are \textbf{trained with only SA-1B} and directly evaluate on COCO \textit{Val2017} for fair comparison. \textit{Max} denotes we select the output with the max confidence score prediction. \textit{Oracle} denotes we select the output with the max IoU by calculating the IoU between the prediction and target mask.}
\label{tab:interactive}
\end{table*}

In Table~\ref{tab:interactive}, we evaluate the 1-click mIoU (denoted as 1-IoU) for SAM and our model on COCO \textit{Val2017}. Our model outperforms SAM under the same settings.

\paragraph{Multi-granularity Interactive Segmentation}
In Table~\ref{tab:multi-granularity}, we compare SAM~\cite{kirillov2023segment} and our model on the output granularities for a single click. We adopt a Hungarian Matching to match all the possible target masks with the predicted masks for the click and calculate the average IoU score. As SAM has only three prompts, we also sample two clicks from a single mask to produce six output masks for a fair comparison. Notably, SAM has been trained on this validation set while we did not.
\begin{table*}
\centering
\begin{tabular}{lc|ccc|cc} 
\hline
{Method} &Granularity& 1-IoU@All Granularity \\
\hline
SAM (B)$^\dag$ &3& 75.6\\
SAM (L)$^\dag$ &3& 82.5\\
SAM (H)$^\dag$ &3& 83.5\\
SAM (B)$^\dag$$^*$ &6& 79.3\\
SAM (L)$^\dag$$^*$ &6& 85.6\\
SAM (H)$^\dag$$^*$ &6& 86.5\\

\ourmodel{}(T) &6&88.1\\
\ourmodel{}(L) &6&89.0\\
\hline
\end{tabular}
\caption{Granularity comparison between SAM and our model on a subset of SA-1B with 1000 images. We did not train on this subset of images but SAM did. For each click, we evaluate all the possible ground-truth masks to calculate the \textit{1-IoU@All Granularity}. SAM~\cite{kirillov2023segment} and \ourmodel{} adopts three and six prompts for a single click of a mask, respectively. $^\dag$ denotes that \textit{SAM has been trained on this validation subset while we did not}. $^*$ denotes that we click two points for a single mask to produce six output masks.}
\label{tab:multi-granularity}
\end{table*}
\subsection{Abaltions}
\paragraph{Match Strategy}


\begin{table*}
\centering
\begin{tabular}{lc|ccc|cc} 
\hline
{Method} &Match& 1-IoU@All Granularity \\
\hline
\ourmodel{}(T) &Many-to-one&{73.2}\\
\ourmodel{}(T) &Many-to-many&\textbf{88.1}\\

\hline
\end{tabular}
\caption{Different match strategy comparison on output granularity.}
\label{tab:match_compare}
\end{table*}

As shown in Table~\ref{tab:match_compare}, we compare different match strategies in our model. When using many-to-many matching to match all the possible ground-truth masks for each click, the 1-IoU@All Granularity performance is significantly improved. This validates our matching strategy is effective to learn complete granularities.
\paragraph{Box Interactive Evaluation}
\begin{table*}
\centering
\begin{tabular}{l|c} 
\hline
{Method}& Box 1-IoU \\
\hline
SAM~\cite{kirillov2023segment}(B) &{50.7}\\
SEEM~\cite{zou2023segment}(T) &{73.7}\\
\ourmodel{}(T) &\textbf{76.1}\\

\hline
\end{tabular}
\caption{Box 1-IoU evaluation on COCO \textit{Val2017}. Both SEEM~\cite{zou2023segment} and our model are trained on COCO and we additionally train on SA-1B~\cite{kirillov2023segment}.}
\label{tab:box_compare}
\end{table*}

We also evaluate the 1-IoU given boxes in Table~\ref{tab:box_compare}. We achieve better performance compared with object-level interactive segmentation model SEEM~\cite{zou2023segment} and multi-granularity model SAM~\cite{kirillov2023segment}.
\paragraph{Increasing SA-1B Training data}

\begin{table*}
\centering
\begin{tabular}{l|c|ccc} 
\hline
\multirow{3}{*}{Method}&\multirow{3}{*}{Data Portion of SA-1B} & \multicolumn{2}{c}{COCO}  \\
&& Point (Max) & Point (Oracle)   \\
&& 1-IoU & 1-IoU \\ 
\hline
SAM (L) &100\% &55.7&70.5\\

\ourmodel{} (L) &3\%&55.2&73.5\\
\ourmodel{} (L) &15\%&56.7&73.6 \\
\ourmodel{} (L) &30\%&55.7&73.7 \\
\ourmodel{} (L) &50\%&55.3&73.9 \\
\ourmodel{} (L) &100\%&57.0&74.2 \\
\hline
\end{tabular}
\caption{Comparison of using different portions of SA-1B~\cite{kirillov2023segment} data. Our model is only trained with SA-1B and directly evaluated on COCO \textit{Val2017}. 
}
\label{tab:data_scale}
\end{table*}
In Table~\ref{tab:data_scale}, we show the performance improvement on COCO \textit{Val 2017} when training with more SA-1B data. The performance is saturated after using more than 15\% of the total data. It indicates that we do not need
to train with the whole SA-1B data to get a good zero-shot performance.

\subsection{Visualization}
\begin{figure}[t!]
    \centering
    \includegraphics[width=0.97\textwidth]{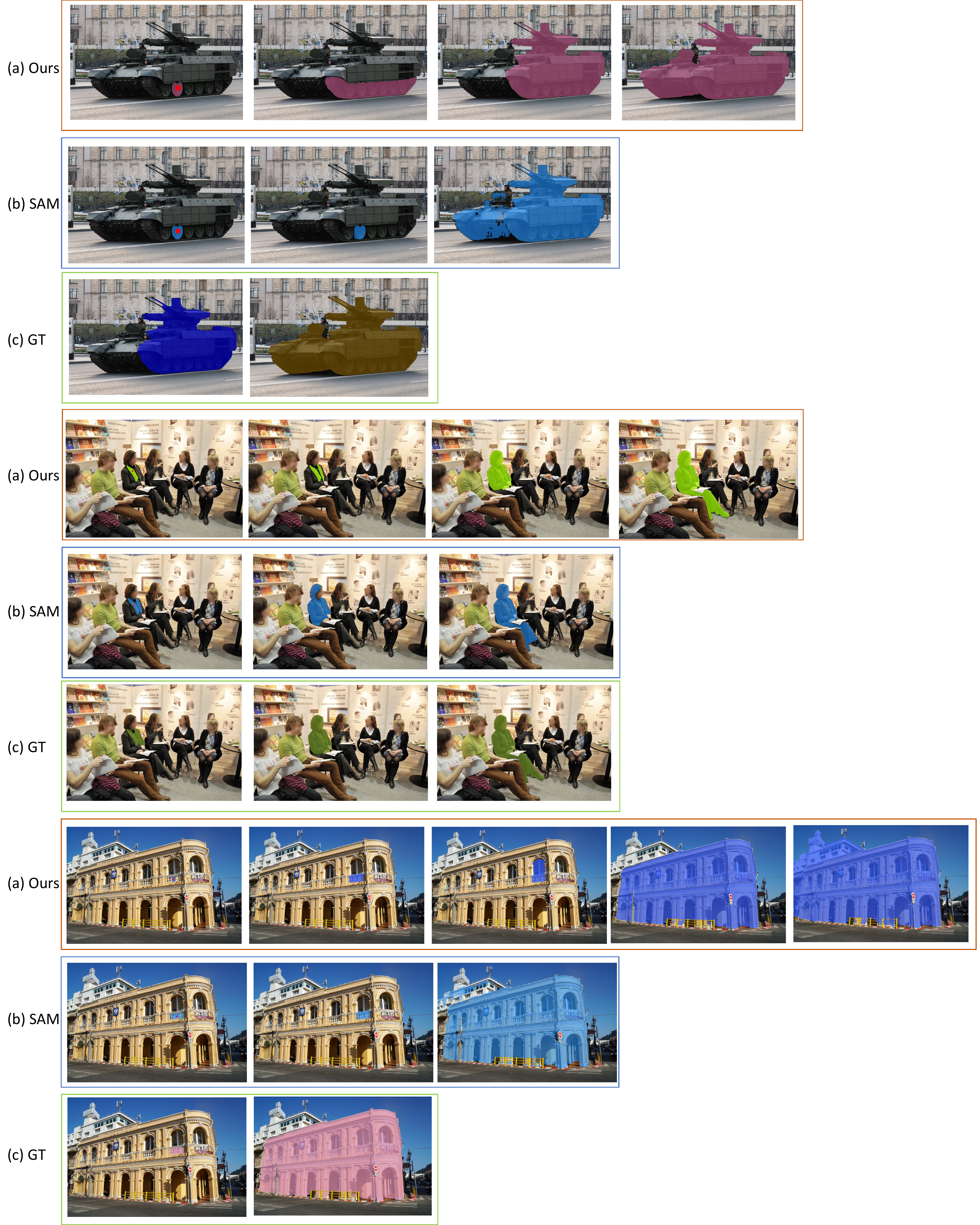}
    \vspace{-5pt}
    \caption{(a)(b) are the output masks of our model and SAM, respectively. The red points on the left-most image of each row are the use clicks. (c) shows the GT masks that contain the user clicks. The outputs of our model have been processed to remove duplicates.}
    \label{fig: compare sam}
    \vspace{-12pt}
\end{figure}
We compare our model with SAM to show that our model can output more levels of high-quality masks, as shown in Fig.~\ref{fig: compare sam}. \\

\noindent\textbf{Multi-Level Masks} Our model outputs more meaningful granularities of masks. SAM outputs three masks at most and different levels of outputs are sometimes duplications, While, the output masks of our model are more diverse.

\noindent\textbf{Mask Qualities}
It is also proved that our model output masks with higher quality. SAM sometimes outputs masks with artifacts such as holes or islands especially for large masks when the click is within a small-scale mask, while our model output high-quality masks for all levels.

\paragraph{Compare with SA-1B Ground-truth Granularity}
We output more meaningful granularity on SAM data compared with the original annotation.

\paragraph{Query semantics}
\begin{figure}[t!]
    \centering
    \includegraphics[width=0.97\textwidth]{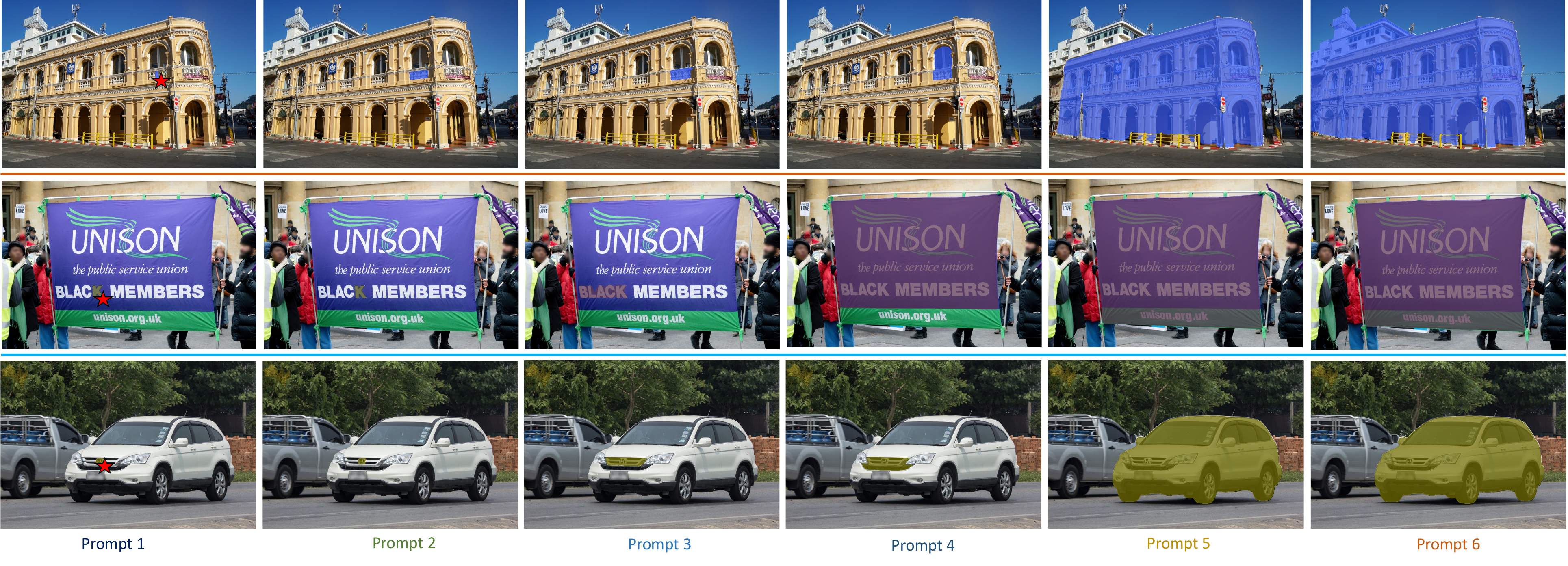}
    \vspace{-5pt}
    \caption{We visualize the prediction of each content prompt embedding of points with a fixed order for our model. We find all the output masks are from small to large. This indicates each prompt embedding represents a semantic level.}
    \label{fig: levels}
    \vspace{-12pt}
\end{figure}
We also find that each point content prompt embeddings learns to correspond to a fixed granularity. As shown in Fig.~\ref{fig: levels}, when we visualize masks in a specific order of the corresponding content embeddings, the masks follow the order from small to large in each row consistently. This proves that each content embedding represents a semantic granularity level in our model.

\section{Related works}

\subsection{Generic Segmentation}  
Segmenting visual concepts is well-documented within the expansive field of computer vision~\cite{fu1981survey,felzenszwalb2009object,zou2019object,minaee2021image}. Broad segmentation methodologies comprise several subdivisions, such as instance segmentation, semantic segmentation, and panoptic segmentation~\cite{he2017mask,chen2017deeplab,kirillov2019panoptic}, each catering to a unique semantic degree. For example, semantic segmentation's goal is to detect and assign a label to each pixel in an image according to its corresponding semantic class~\cite{chen2017rethinking, cheng2022masked, long2015fully}. Conversely, instance segmentation seeks to cluster pixels associated with the same semantic class into distinct object instances~\cite{he2017mask, bolya2019yolact,li2022mask}. Panoptic segmentation is the hybrid of these two tasks.
Recently, Transformer-based methods~\cite{vaswani2017attention,carion2020end} have contributed to significant progress in segmentation tasks~\cite{li2022panoptic, cheng2022masked, li2022mask,jain2022oneformer,zhang2023mp}.
Generic object detection and segmentation have led to the development of abundant datasets, such as MSCOCO~\cite{lin2014microsoft}, LVIS~\cite{gupta2019lvis}, Objects365~\cite{shao2019objects365}, PASCAL~\cite{everingham2011pascal},CityScapes~\cite{cordts2016cityscapes},ADE20k~\cite{zhou2018semantic}, etc. 

\subsection{Part Segmentation}
Beyond generic segmentation, part segmentation aims to more fine-grained visual understanding.
Most early works were bottom-up methods by grouping super-pixels into parts and then objects ~\cite{arbelaez2010contour,grundmann2010efficient,arbelaez2014multiscale}. Later, based on high-performance object detection networks~\cite{ren2015faster,he2017mask}, top-down methods were developed by firstly 
detecting an object and then parsing it to part segmentation~\cite{li2017holistic,yang2019parsing,ji2020learning}. To segment the scene in multi-granularity, part-aware panoptic segmentation~\cite{de2021part} is introduced.  PPS~\cite{de2021part} establishes the baseline through assembling panoptic and part segmentation models. JPPF~\cite{jagadeesh2022multi} simplifies the model by a shared image encoder for both panoptic segmentation and part segmentation. By representing thing, stuffs, and parts as object queries, Panoptic-PartFormer~\cite{li2022panopticpart} proposes a unified architecture based on Transformer. While part segmentation data is much expensive than
object detection and segmentation data, a number of public datasets are available. Datasets for specific domains include cars~\cite{song2019apollocar3d}, birds~\cite{wah2011caltech}, and fashion~\cite{jia2020fashionpedia}. General objects include Pascal-Part~\cite{chen2014detect}, PartImageNet~\cite{he2021partimagenet}, ADE20K~\cite{zhou2017scene}, Cityscapes-Panoptic-Parts~\cite{meletis2020cityscapes}, and PACO~\cite{ramanathan2023paco}. More recently, SAM~\cite{kirillov2023segment} provides a large-scale multi-granularity class-agnostic segmentation dataset. Our work is jointly trained on these datasets and contributes to a multi-granularity segmentation model.

\subsection{Open-Vocabulary Segmentation} 
While generic segmentation and part segmentation have made remarkable progress, they can only segment the image in a close-set vocabulary. To expand the vocabulary size, recent works leverage the visual-semantic knowledge from large-scale foundation models like CLIP~\cite{radford2021learning}, ALIGN~\cite{jia2021scaling} and Diffusion models~\cite{xu2023open} to various segmentation tasks. LSeg~\cite{li2022language}, OpenSeg~\cite{ghiasi2021open}, GroupViT~\cite{xu2022groupvit} achieves open-vocabulary semantic segmentation ability on ADE20K and PASCAL. DenseCLIP~\cite{rao2022denseclip} and MaskCLIP~\cite{ding2022open} achieves open-vocabulary  instance and panoptic segmentation on COCO dataset. More recently, X-Decoder~\cite{zou2022generalized} proposes a unified approach to tackle various segmentation and vision-language tasks for open-vocabulary segmentation, OpenSeeD~\cite{zhang2023simple} proposes to use a large amount of detection data and a joint training method to improve segmentation. To segment open-vocabulary masks in part-level, VLPart~\cite{sun2023going} leverages three part segmentation datasets and learns from the dense correspondence~\cite{caron2021emerging} between base objects and novel objects. Our work unifies these tasks into one architecture and builds up open-vocabulary segmentation in multi-granularity.

\subsection{Interactive Segmentation}
Interactive segmentation refers to the process of separating objects by actively integrating user inputs. This enduring challenge has seen notable advancements~\cite{li2004lazy, grady2006random, xu2016deep, liu2022simpleclick, chen2022focalclick, kirillov2023segment}. Previous works only focus on a small set of data or semantic-agnostic instance masks. Recently, SAM~\cite{kirillov2023segment} enlarges the training data from 0.12M COCO images to 10M SAM fine-grained images. And SEEM~\cite{zou2023segment} enriches the modality to language and function to both generic and grounded segmentation with an impressive compositionality. 

\section{Conclusion}
In this paper, we have presented \textit{\ourmodel}, which can segment and recognize anything at any desired granularity. Apart from performing generic open-vocabulary segmentation, \ourmodel{} demonstrates the advantages of semantic awareness and granularity abundance. To achieve such advantages, we have proposed improvements on data, model, and training where we utilized datasets from multiple granularity and semantic levels, multi-choice learning for training, and a universal framework for modeling. Comprehensive experiments and visualizations have verified the semantic awareness and granularity abundance of our model. Further, \ourmodel{} is the first successful attempt to jointly train on SA-1B and other classic segmentation datasets. Experimental results also show that training with SA-1B improves other tasks such as panoptic and part segmentation.

\bibliographystyle{ieee_fullname}
\bibliography{egbib}

\end{document}